\documentclass[journal]{IEEEtran}
\IEEEoverridecommandlockouts
\usepackage{soul}
\usepackage{cite}
\usepackage{amsmath,amssymb,amsfonts}
\usepackage{algorithmic}
\usepackage{graphicx}
\usepackage{textcomp}
\usepackage{xcolor}

\usepackage[left=0.68in, right=0.68in, bottom=0.7in, top=0.772in]{geometry}

\colorlet{merim}{blue}

\usepackage[caption=false,labelformat=simple]{subfig}

\def\BibTeX{{\rm B\kern-.05em{\sc i\kern-.025em b}\kern-.08em
    T\kern-.1667em\lower.7ex\hbox{E}\kern-.125emX}}

\usepackage{pgfplots}
\pgfplotsset{compat=1.5}
\usepackage{pgfplotstable}
\usepgfplotslibrary{statistics}
\pgfplotsset{grid style={dotted,gray}}
\usepackage{tikz}
\usepackage[toc,acronym]{glossaries}
\newacronym{3gpp}{3GPP}{Third Generation Partnership Project}
\newacronym{ran}{RAN}{Radio Access Network}
\newacronym{oran}{O-RAN}{Open Radio Access Network}
\newacronym{ml}{ML}{Machine Learning}
\newacronym{ai}{AI}{Artificial Intelligence}
\newacronym{sba}{SBA}{Service-based Architecture}
\newacronym{5g}{5G}{Fifth-Generation}
\newacronym{4g}{4G}{Fourth-Generation}
\newacronym{cn}{CN}{Core Network}
\newacronym{ngran}{NG-RAN}{Next Generation Radio Access Network}
\newacronym{urllc}{URLLC}{Ultra-Reliable Low-Latency Communications}
\newacronym{nf}{NF}{Network Function}
\newacronym{nef}{NEF}{Network Exposure Function}
\newacronym{nfv}{NFV}{Network Functions Virtualization}
\newacronym{sdn}{SDN}{Software-Defined Networking}
\newacronym{sg}{SG}{Smart Grid}
\newacronym{pcf}{PCF}{Policy Control Function}
\newacronym{iot}{IoT}{Internet of Things}
\newacronym{se}{SE}{State Estimation}
\newacronym{ems}{EMS}{Energy Management System}
\newacronym{wls}{WLS}{Weighted Least Squares}
\newacronym{ami}{AMI}{Advanced Metering Infrastructure}
\newacronym{wampac}{WAMPAC}{Wide-Area Monitoring, Protection, and Control}
\newacronym{wams}{WAMS}{Wide-Area Monitoring System}
\newacronym{mmtc}{mMTC}{Massive Machine-type Communications}
\newacronym{pmu}{PMU}{Phasor Measurement Unit}
\newacronym{gbp}{GBP}{Gaussian Belief Propagation}
\newacronym{gnn}{GNN}{Graph Neural Networks}
\newacronym{mdaf}{MDAF}{Management Data Analytics Function}
\newacronym{nwdaf}{NWDAF}{Network Data Analytics Function}
\newacronym{af}{AF}{Application Function}
\newacronym{pdc}{PDC}{Phasor Data Concentrators}
\newacronym{dso}{DSO}{Distribution System Operator}
\newacronym{ied}{IED}{Intelligent Electronic Device}
\newacronym{nr}{NR}{New Radio}
\newacronym{ue}{UE}{User Equipment}
\newacronym{upf}{UPF}{User Plane Function}

\pgfplotsset{legend image with text/.style={legend image code/.code={%
\node[anchor=west, align=right] at (0.0cm,0cm) {#1};}},}

\pgfplotsset{
    box plot/.style={
        /pgfplots/.cd,
        black,
        only marks,
        mark=-,
        mark size=\pgfkeysvalueof{/pgfplots/box plot width},
        /pgfplots/error bars/y dir=plus,
        /pgfplots/error bars/y explicit,
        /pgfplots/table/x index=\pgfkeysvalueof{/pgfplots/box plot x index},
    },
    box plot box/.style={
        /pgfplots/error bars/draw error bar/.code 2 args={%
            \draw  ##1 -- ++(\pgfkeysvalueof{/pgfplots/box plot width},0pt) |- ##2 -- ++(-\pgfkeysvalueof{/pgfplots/box plot width},0pt) |- ##1 -- cycle;
        },
        /pgfplots/table/.cd,
        y index=\pgfkeysvalueof{/pgfplots/box plot box top index},
        y error expr={
            \thisrowno{\pgfkeysvalueof{/pgfplots/box plot box bottom index}}
            - \thisrowno{\pgfkeysvalueof{/pgfplots/box plot box top index}}
        },
        /pgfplots/box plot
    },
    box plot top whisker/.style={
        /pgfplots/error bars/draw error bar/.code 2 args={%
            \pgfkeysgetvalue{/pgfplots/error bars/error mark}%
            {\pgfplotserrorbarsmark}%
            \pgfkeysgetvalue{/pgfplots/error bars/error mark options}%
            {\pgfplotserrorbarsmarkopts}%
            \path ##1 -- ##2;
        },
        /pgfplots/table/.cd,
        y index=\pgfkeysvalueof{/pgfplots/box plot whisker top index},
        y error expr={
            \thisrowno{\pgfkeysvalueof{/pgfplots/box plot box top index}}
            - \thisrowno{\pgfkeysvalueof{/pgfplots/box plot whisker top index}}
        },
        /pgfplots/box plot
    },
    box plot bottom whisker/.style={
        /pgfplots/error bars/draw error bar/.code 2 args={%
            \pgfkeysgetvalue{/pgfplots/error bars/error mark}%
            {\pgfplotserrorbarsmark}%
            \pgfkeysgetvalue{/pgfplots/error bars/error mark options}%
            {\pgfplotserrorbarsmarkopts}%
            \path ##1 -- ##2;
        },
        /pgfplots/table/.cd,
        y index=\pgfkeysvalueof{/pgfplots/box plot whisker bottom index},
        y error expr={
            \thisrowno{\pgfkeysvalueof{/pgfplots/box plot box bottom index}}
            - \thisrowno{\pgfkeysvalueof{/pgfplots/box plot whisker bottom index}}
        },
        /pgfplots/box plot
    },
    box plot median/.style={
        /pgfplots/box plot,
        /pgfplots/table/y index=\pgfkeysvalueof{/pgfplots/box plot median index},
        semithick,black
    },
    box plot width/.initial=1em,
    box plot x index/.initial=0,
    box plot median index/.initial=1,
    box plot box top index/.initial=2,
    box plot box bottom index/.initial=3,
    box plot whisker top index/.initial=4,
    box plot whisker bottom index/.initial=5,
}

\newcommand{\boxplot}[2][]{
    \addplot [box plot median,#1] table {#2};
    \addplot [forget plot, box plot box,#1] table {#2};
    \addplot [forget plot, box plot top whisker,#1] table {#2};
    \addplot [forget plot, box plot bottom whisker,#1] table {#2};
}

\begin{document}

\title{Near Real-Time Distributed State Estimation via AI/ML-Empowered 5G Networks}

\author{Ognjen Kundacina,
        Miodrag Forcan,
        Mirsad Cosovic,
        Darijo Raca,
        Merim Dzaferagic,
        Dragisa Miskovic,
        Mirjana Maksimovic,
        Dejan Vukobratovic

\thanks{O. Kundacina and D. Miskovic are with The Institute for Artificial Intelligence Research and Development of Serbia (e-mail: ognjen.kundacina@ivi.ac.rs, dragisa.miskovic@ivi.ac.rs); 
M. Forcan and M. Maksimovic are with Faculty of Electrical Engineering, University of East Sarajevo, Bosnia and Herzegovina (e-mail: miodrag.forcan@etf.ues.rs.ba, mirjana.maksimovic@etf.ues.rs.ba); M. Cosovic and D. Raca are with Faculty of Electrical Engineering, University of Sarajevo, Bosnia and Herzegovina (e-mail: mcosovic@etf.unsa.ba, draca@etf.unsa.ba); M. Dzaferagic is with CONNECT centre, Trinity College Dublin, Ireland (e-mail: dzaferam@tcd.ie); D. Vukobratovic is with Faculty of Technical Sciences, University of Novi Sad, Serbia, (email: dejanv@uns.ac.rs).}
\thanks{This paper has received funding from the European Union's Horizon 2020 research and innovation programme under Grant Agreement number 856967.}}

\maketitle

\begin{abstract}
\gls{5g} networks have a potential to accelerate power system transition to a flexible, softwarized, data-driven, and intelligent grid. With their evolving support for \gls{ml}/\gls{ai} functions, \gls{5g} networks are expected to enable novel data-centric \gls{sg} services. In this paper, we explore how data-driven \gls{sg} services could be integrated with \gls{ml}/\gls{ai}-enabled \gls{5g} networks in a symbiotic relationship. We focus on the \gls{se} function as a key element of the energy management system and focus on two main questions. Firstly, in a tutorial fashion, we present an overview on how distributed \gls{se} can be integrated with the elements of the \gls{5g} core network and radio access network architecture. Secondly, we present and compare two powerful distributed \gls{se} methods based on: i) graphical models and belief propagation, and ii) graph neural networks. We discuss their performance and capability to support a near real-time distributed \gls{se} via \gls{5g} network, taking into account communication delays.  
\end{abstract}

\glsresetall
\section{Introduction}
\gls{5g} mobile cellular networks are evolving towards a ubiquitous platform that offers large-scale and distributed communication and computation capacities to support future \gls{ml} and \gls{ai}-based services \cite{li2017intelligent}. Integration of \gls{sg} with \gls{5g} network will foster accelerated transition to a flexible, softwarized, data-driven and intelligent grid, recognised by the \gls{3gpp} work plan for Release 18 of \gls{5g} standards \cite{xia2021review}. \gls{5g} networks provide necessary bandwidth, connection density, low latency and ultra-reliability to support different \gls{sg} services \cite{dragivcevic2019future}. Additionally, \gls{5g} networks are continuously upgraded to offer an increasing in-network support for \gls{ml}/\gls{ai}-enabled services \cite{chouman2022towards}. 

\gls{ami} and \gls{wampac} systems are the major \gls{sg} services that could match well with the emerging \gls{ml}/\gls{ai}-enabled \gls{5g} network services. \gls{ami} demands extreme connection densities ($10^6$ devices/km$^2$) to support smart meter connectivity which is achievable using \gls{5g} \gls{mmtc} \cite{mohassel2014survey, song2019cellular}. \gls{wams}, as a part of \gls{wampac}, delivers \gls{pmu} measurements to the control center for the purpose of \gls{se}, power system monitoring, warning, and event analysis \cite{ekanayake2012smart}. Given the number of connected \gls{ami} devices, and \gls{pmu} reporting rates of $10 - 100$ frames per second, \gls{5g} is capable of supporting extreme data volumes to enable next-generation data-driven \gls{sg} services \cite{bose2017artificial}. Furthermore, relying on \gls{5g} \gls{urllc}, \gls{wams} based on \glspl{pmu} will offer a near real-time view of power system dynamics leading to novel \gls{ml}/\gls{ai}-based monitoring, control and protection services \cite{phadke2016improving}. 

In this paper, we firstly review an ongoing \gls{5g} network evolution of both \gls{5g} \gls{cn} and \gls{5g} \gls{ran} to support \gls{ml}/\gls{ai}-based services. Then, we consider how data-driven \gls{sg} services could be integrated with \gls{ml}/\gls{ai}-enabled \gls{5g} networks in a symbiotic relationship. Our focus is on the \gls{se} as one of the main routines in the \gls{ems} \cite{monticelli2000electric}. In particular, we consider the distributed \gls{se} that exploits distributed computation and communication in order to obtain the global state of the system \cite{korres2010distributed, xie2012fully}. Integrated with \gls{5g} network, the distributed \gls{se} depends on massive and timely local data inputs via \gls{5g} \gls{ran}, and the design and deployment of near real-time distributed algorithms on top of \gls{5g} edge and cloud computing capacities \cite{cosovic20175g}. Herein, we present two powerful distributed \gls{se} methods based on: i) factor graphs and \gls{gbp}, and ii) \gls{gnn}. The performance and latency of a distributed \gls{se} that relies on the message passing between edge computing nodes are investigated. For both model-based methods, that apply GBP approach, and data-based methods, that rely on \gls{gnn} approach, we present their performance, discuss their integration into \gls{ml}/\gls{ai}-enabled \gls{5g} network architecture, and outline their advantages and drawbacks.

\section{5G Support for Smart Grid Services}

In this section, we consider \gls{5g} network support for \gls{ml}/\gls{ai}-based third-party services (e.g., \gls{sg} services) in both centralized cloud-based \gls{5g} \gls{cn} and distributed edge-based \gls{5g} \gls{ran}.    

\begin{figure*}
    \centerline{\includegraphics[width=6.5in]{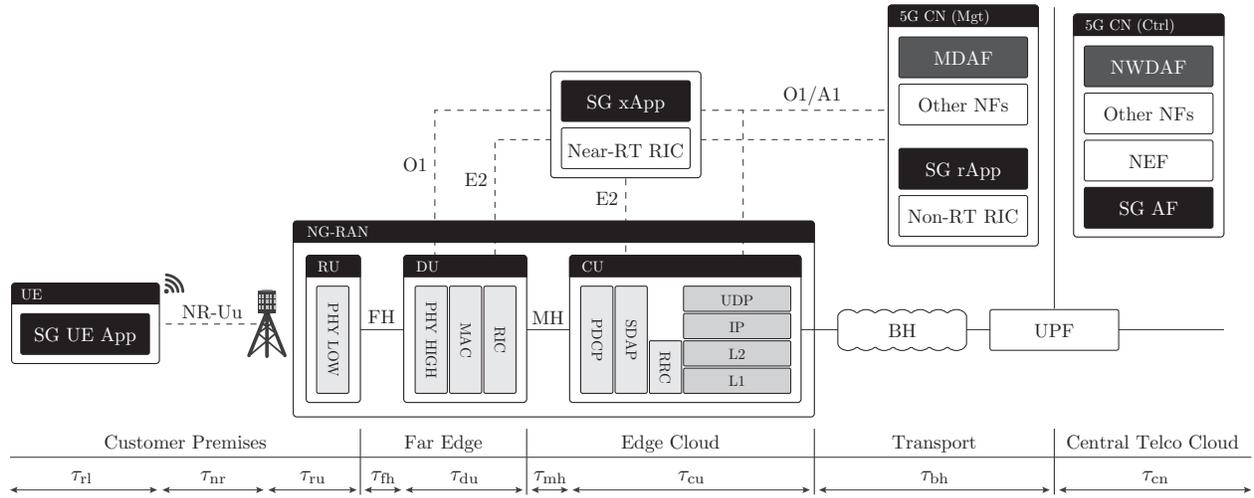}}
    \caption{5G network architecture support for SG services.}
    \label{Fig_1}
\end{figure*}

\subsection{5G Core Network Service-Based Architecture}

Evolution and deployment of novel services with ever increasing requirements motivated \gls{3gpp} to adopt a new \gls{5g} \gls{cn} system architecture called \gls{sba}. Fast and flexible service deployment in \gls{5g} \gls{sba} is critical for commercial viability of \gls{5g} \cite{Lu20195GES, 8788396}. \gls{5g} \gls{sba} is defined as a set of interconnected \glspl{nf} that have permission to access each other's services \cite{3GPP23.501}. Becoming \emph{cloud-native} means that the \gls{5g} \gls{cn} employs \gls{nfv} and \gls{sdn} approaches, and service-based interactions between control-plane functions \cite{borgaonkar20195g,8902883}.

The \gls{5g} \gls{cn} architecture consists of a number of \glspl{nf} responsible for network control and management, as shown in \figurename~\ref{Fig_1} \cite{3GPP23.501,8788396}. The \gls{cn} design principles include: separation of the control and user planes, modularization of the functional and interface design, support for direct interaction of \glspl{nf}, reduction of mutual reliance of the \gls{5g} \gls{ran} and \gls{cn}, support for \emph{stateless} \glspl{nf} by separating computing and storage resources, exposure of capability, and allowance of concurrent access to both local and centralized services \cite{3GPP23.501,nguyen2017sdn,7993854}. Besides standard \glspl{nf} \cite{nguyen2017sdn}, novel data analytics functions: \gls{mdaf} and \gls{nwdaf}, have been introduced. \gls{mdaf} manages data analytics for one or more \glspl{nf} while \gls{nwdaf} collects and analyzes data for both centralized and edge computing resources. \gls{nwdaf} simplifies the production and consumption of \gls{cn} data, and provides insights and take actions to improve the end-user experience and network performance \cite{3GPP23.501,8788396}.

\glspl{af} communicate securely with other \gls{5g} \glspl{nf} via the \gls{5g} \gls{sba}. \glspl{af} can be created for a variety of application services, e.g., Smart Grid AFs (SGAFs), and owned by the network operator or trusted third parties. For services considered trusted by the operator, the \glspl{af} can access \glspl{nf} directly, while untrusted or third-party \glspl{af} access \glspl{nf} through the \gls{nef}. \gls{nef} acts as a communication mediator with the external systems and its role is to store and retrieve \glspl{nf} exposed capabilities and events. In \gls{5g} \gls{cn}, this data is exchanged among \glspl{nf} via \gls{nef}, but it can be securely exposed to third-party and edge-computing entities outside of the 3GPP network, e.g., for the purpose of data analytics \cite{3GPP23.501,8788396}.

\begin{figure}
    \centerline{\includegraphics[width=3.4in]{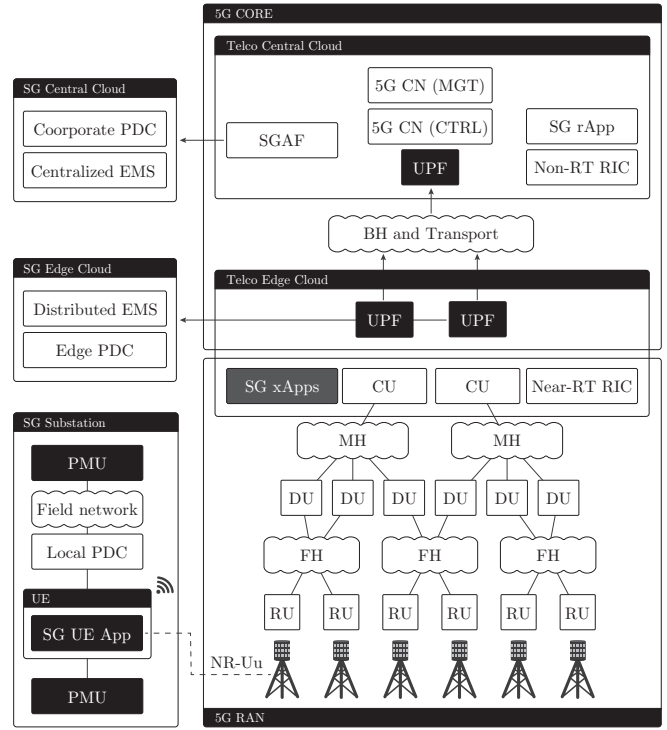}}
    \caption{Example of SG service deployment over 5G network architectures.}
    \label{Fig_2}
\end{figure}
 
\subsection{5G Open Radio Access Network Architecture}

The \gls{5g} \gls{ran} is undergoing the process of \gls{ran} disaggregation, i.e., a transition from inflexible, monolithic and hardware-based \gls{ran} to a flexible, elastic, software-based and disaggregated \gls{ran}. The goal is to optimize radio resource management and efficient use of spectrum. The placement of \gls{ran} functions across different locations started in \gls{4g} in the context of Cloud-\gls{ran} (C-RAN). Base stations (also called eNBs) are separated into BaseBand Units (BBUs) located in central offices and separated from Remote Radio Units (RRUs) \cite{checko2014cloud}. Further disaggregation divides the BBU and the RRU into the Centralised Unit (CU), the Distributed Unit (DU), and the Radio Unit (RU). Such a disaggregation, adopted by the \gls{oran} alliance, is under investigation within \gls{3gpp} targeting the most efficient functional split, i.e., the optimal allocation of RAN functions to CU/DU/RU \cite{3GPP38.801}, see \figurename~\ref{Fig_1}.

The introduction of \gls{ml}/\gls{ai}-based services is gradually expanding from \gls{5g} \gls{cn} towards \gls{5g} RAN. \gls{3gpp} is investigating possible locations for data collection, training and deployment of inference models \cite{3GPP37.817}. In \gls{oran}, novel hierarchical RAN Intelligent Controllers (RIC) are introduced: i) Near-Real-Time (Near-RT) RIC deployed at the edge, and ii) Non-Real-Time (Non-RT) RIC deployed at the central cloud, connected via a new set of open interfaces (E2/O1/A1). Near-RT RIC supports a number of xApps operating at $10\,\text{ms}$ - $1\,\text{s}$ time scales: microservices capable of performing various \gls{ml}/\gls{ai}-based resource management functions (e.g., network slicing, radio resource management). In contrast, rApps have more information about the network traffic (understand the context), thus they are mainly used for making predictions and submitting policies to different network nodes. They also share policies with the near-RT RIC. Then, those policies are being enforced by the near-RT control loop. The near-RT RIC can also collect data from multiple CUs and DUs and make coordinated control decisions for multiple nodes \cite{polese2022understanding}. Although xApps/rApps address RAN management and control, third-party xApps/rApps will enable integration and control of specific network service requirements, e.g., \gls{sg} applications.        






\gls{5g} \gls{urllc} services are suitable for \gls{pmu}-based \gls{wams} in \gls{sg} that target near real-time system state awareness. A \gls{urllc}-based distributed \gls{se} for \gls{sg} with latency and reliability considerations is proposed in \cite{cosovic20175g}. Similarly, the implementation of \gls{se} functions in \gls{sg} using \gls{5g} was recently introduced by \cite{zerihun2020effect}. The challenges and promises of \gls{urllc} for \gls{sg} teleprotection are elaborated in \cite{ghanem2021challenges}. In \cite{hovila20195g}, a wireless line differential protection scheme based on \gls{urllc} is proposed and tested using the real \gls{5g} test network. A testbed composed of a microgrid and \gls{5g} network is introduced in \cite{demidov2022iec} and used to evaluate IEC-61850 \cite{international2013communication} protocols in the case of TCP/UDP message transmission. \gls{5g} network slicing for \gls{sg} is implemented and verified in \cite{sun2021implementation}. Unlike the above papers, we focus on \gls{ml}/\gls{ai}-based distributed \gls{se} methods and their synergies with \gls{ml}/\gls{ai}-empowered \gls{5g} network.

\section{Distributed State Estimation Architecture in ML/AI-Empowered 5G Networks}


\subsection{PMUs: Near Real-Time Data Sources for 5G Network}

\gls{wams} relies on \glspl{pmu} that generate synchronized voltage and current phasor measurements \cite{IEEEIECInternationalStandard}. \glspl{pmu} can be viewed as sensors used to perform wide-area phasor measurements in power systems \cite{liu2017distribution}. \glspl{pmu} are designed to support multiple data reporting rates, usually up to $100$ frames per second in the case of $50\,\text{Hz}$ power systems. In addition to \glspl{pmu}, \gls{pdc} are used in communication systems in order to correlate synchrophasor data by time stamps creating a system wide measurement sets \cite{IEEEStandardforSynchrophasor}.

Depending on \gls{sg} architecture, both \glspl{pmu} and \glspl{pdc} can access the network as a \gls{5g} \gls{nr} \gls{ue}. \glspl{pmu} can be deployed in two scenarios: i) as stand-alone devices that send data as UEs to a remote \gls{pdc}, or ii) as part of a local field network deployed by a power system operator that connects multiple \glspl{pmu} in a given region to a \gls{pdc}, which in turn connects to the \gls{5g} network as a \gls{ue} (see \gls{sg} Substation block in \figurename~\ref{Fig_2}). Besides local \glspl{pdc}, power system operator may deploy regional or corporate \glspl{pdc} envisioned as edge-cloud or central-cloud applications (\figurename~\ref{Fig_2}). Edge \glspl{pdc} are usually deployed by a \gls{dso} as part of the distribution network \gls{se}. Corporate \glspl{pdc} usually provide data to a single utility (state-level transmission network), while regional \glspl{pdc} provide data at interconnection level (multi state-level transmission network).

\begin{figure}
    \centering
    \includegraphics[width=3.3in]{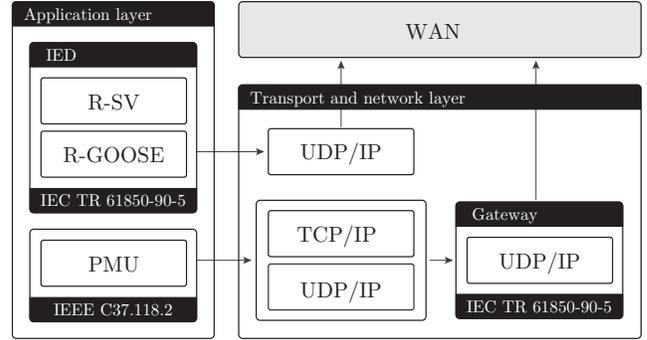}
    \caption{OSI layers related to PMU wide-area communications, and compliant with IEEE C37.118.2 and IEC TR 61850-90-5 protocols.}
    \label{Fig_3}
\end{figure}

Information stored in \gls{pmu}/\gls{pdc} should be formatted as defined in IEEE C37.118.2 \cite{IEEEStandardforSynchrophasor}. The most widely used protocols for the transfer of synchrophasor data are IEEE C37.118.2 \cite{IEEEStandardforSynchrophasor} and IEC TR 61850-90-5 \cite{IECcommunicationnetworks}. IEC TR 61850-90-5 protocol bridges the gap between IEC 61850, recommended for \glspl{ied} at electrical substations \cite{international2013communication}, and IEEE C37.118.2 designed specifically for \glspl{pmu}. It enables wide-area synchrophasor data transfer using Routed Sampled Value (R-SV) and Routed Generic Object-Oriented Substation Event (R-GOOSE) protocols. Application, transport, and network layer encapsulation related to \gls{pmu} communications compliant with IEEE C37.118.2 and IEC TR 61850-90-5 protocols, are shown in \figurename~\ref{Fig_3}. Upon UDP/IP encapsulation of \gls{pmu}/\gls{pdc} data frames, they are transmitted by 5G \gls{ue} using \gls{urllc} service \cite{demidov2022iec}.

\subsection{Distributed SE in ML/AI-Empowered 5G Networks}
The \gls{se} is a key functionality of the \gls{ems} whose aim is to provide a timely estimate of the system state variables (magnitude and angle of the voltage) at all the buses of the power system \cite{cosovic20175g}. Traditional \gls{ems} is centralised, where data collection and its processing is concentrated at a single node. \gls{5g} supports centralised \gls{ems} by deploying its functions as cloud-native applications within the \gls{sg} central cloud (see Centralized \gls{ems} module in \figurename~\ref{Fig_2}). Centralized \gls{ems} runs centralised \gls{se} algorithm based on the set of \gls{pmu} measurements collected at the \gls{5g} central cloud either from local or regional (edge) \glspl{pdc}. Centralized \gls{ems} may interact with the \gls{5g} \gls{cn} through SGAF deployed in the \gls{5g} central cloud. SGAF may expose its own data and access relevant network statistics exposed by other \gls{5g} \gls{cn} data analytics functions such as \gls{mdaf} or \gls{nwdaf} in order to improve \gls{ems} services. 

Recent trends see shifting the \gls{se} functionality from centralised to distributed system architecture. 
Distributed \gls{se} implies the absence of the central coordinator, where each local area communicates only with its neighbors \cite{caro2011decentralized, cosovic2018distributed, kekatos2012distributed, matamoros2016multiarea}. In terms of estimation accuracy, the distributed approach is equivalent to the hierarchical and centralized. For latency-critical scenarios, distributed \gls{se} over \gls{5g} networks with distributed information acquisition and processing represents the most promising approach. In the envisioned architecture, the distributed \gls{se} could be virtualized and deployed in numerous \gls{sg} edge cloud servers across the network (see Distributed \gls{ems} module in \figurename~\ref{Fig_2}). Distributed \gls{se} modules receive local \gls{pmu} measurements via \gls{5g} \gls{ran} either directly from \gls{5g}-connected \glspl{pmu}, from \gls{5g}-connected local \glspl{pdc}, or from edge \glspl{pdc}. Based on the received measurements, distributed \gls{se} performs message-passing among neighbouring \gls{se} modules, by exchanging UDP/IP packets through \gls{5g} \gls{cn}. The following subsection provides more details on emerging distributed \gls{se} algorithms based on message-passing, while their performance is numerically evaluated in Section \ref{Sec:results}.

\subsection{Model-Based versus Data-Based Distributed SE} \label{DistSE}

\subsubsection{Model-Based Distributed SE}
The state-of-the-art model-based distributed \gls{se} algorithms exploit the matrix decomposition techniques applied over the \gls{wls} method. In particular, the \gls{se} algorithms based on distributed optimization combined with the alternating direction method of multipliers (ADMM) have become popular in the literature. To decentralise an optimisation problem, ADMM decouples the objective function with consensus variables. The resulting algorithm can be interpreted as an iterative message-passing procedure, in which agents solve subproblems independently~\cite{cosovic20175g}. 

Another efficient iterative message-passing algorithm for distributed inference is \gls{gbp}. Therein, the power system network with a given measurement configuration is mapped onto an equivalent factor graph containing the set of factor
and variable nodes. Factor nodes are defined by the set of measurements, measurement error and measurement function. The variable nodes are determined by the set of state variables. When applied on factor graphs, the \gls{gbp} algorithm calculates the marginal distributions of the system of random variables~\cite{cosovic20175g, cosovic2018distributed}. 

\subsubsection{Data-Based Distributed SE}
The growing collection of historical measurement data in juxtaposition with complex, often untractable problems has increased interest in developing data-driven \gls{se} methods. Data-driven \gls{se}, unlike model-based methods, can be designed to avoid using power system's parameters if they are highly uncertain \cite{antona2018PSUncertainty}. Deep learning-based approaches that are completely data-driven or hybrid, are typically designed using feed-forward or recurrent neural networks \cite{zhang2019}\cite{zamzam2019} which must be trained on sets of samples with a fixed power system topology, and do not offer a possibility of distributed implementation. 

Recent advancements in \glspl{gnn} \cite{geomDeepLearning} solve the problems specific to applying deep learning in power systems. \glspl{gnn} learn from the graph-structured data by recursively aggregating the neighboring node vector embeddings, and transforming them non-linearly into the new embedding space. The node embeddings are initialised by dataset inputs, followed by a predefined number of neighborhood aggregations $k$, the \gls{gnn} outputs the final node embeddings that can be used for classification or regression problems. Apart from not being restricted to training and test examples with fixed topologies, \glspl{gnn} have fewer trainable parameters, lower memory requirements, and can easily incorporate connectivity information into the learning process. 

In recent proposals for \gls{gnn}-based \gls{se} \cite{TGCN_SE_2021} \cite{kundacina2022state}, the \gls{gnn} models are trained to predict state variables based on the dataset of power system's measurements annotated with the node voltage values. The centralised implementation of the trained \gls{gnn} model's inference results in linear computational complexity with the number of nodes in the power system (assuming the constant node degree). Additionally, \gls{gnn}-based \gls{se} can be computationally and geographically distributed across multiple processing units, with the requirement that all of the measurements in the $k$-hop neighbourhood are gathered and sent into the unit that predicts the state variables for each node. Furthermore, the study \cite{kundacina2022state} examines the robustness of the \gls{gnn}-based \gls{se} to communication failures or \gls{pmu} malfunctions that make the \gls{se} problem unobservable, demonstrating that result deterioration occurs only in the neighbourhood of the lost measurements.

\section{Numerical results and discussion} \label{Sec:results}
In this section, we explore the \gls{se} supported by \gls{wams} using the IEEE 30-bus and IEEE 118-bus test case. \glspl{pmu} can accurately measure voltage and current phasors and send data at high reporting rates, requiring efficient algorithms with minimal computational latency to process their measurements. We select two linear computational complexity methods for comparison: the model-based \gls{gbp} method and the data-based \gls{gnn} method.

\subsection{Accuracy of GBP and GNN-based SE}
In all simulation models, we start with a given IEEE test case and apply power flow analysis to generate exact solutions. Furthermore, we induce errors in exact solutions of the magnitudes and angles of the branch currents and bus voltages by applying the additive white Gaussian noise of variance $v = 10^{-5}$ to exact values. The resulting set of measurements represents the \gls{pmu} data source for the \gls{se} model. In particular, each \gls{pmu} placed at a given bus corresponds to the bus voltage phasor and current phasor measurements along branches incident to the bus. Using the optimal placement algorithm given in \cite{gou}, we observe IEEE 30-bus and 118-bus power systems with 10 and 32 \glspl{pmu}, respectively. 

We compare the behaviour of the \gls{gbp} and \gls{gnn} algorithms in the dynamic scenario, where power systems change load values in discrete time instances $\tau = \{1,2,\dots,100\}$, described with 100 different measurement sets. To evaluate \gls{se} algorithms, we select the Weighted Residual Sum of Squares (WRSS) as a evaluation metric:
\begin{equation}
	\mathrm{WRSS}^{(\tau)} = \sum_{i = 1}^m \cfrac{\left[z_i^{(\tau)} - h_i(\hat{\mathbf{x}}^{(\tau)}) \right]^2}{v_i^{(\tau)}},
	\label{rmse}
\end{equation}
where $m$ is the number of measurements, $z_i^{(\tau)}$ and $v_i^{(\tau)}$ are the measurement value and variance in the corresponding time instance $\tau$, and $h_i(\hat{\mathbf{x}}^{(\tau)})$ represents the measurement function evaluated at the point defined by the estimate vector $\hat{\mathbf{x}}^{(\tau)}$. Next, we normalize the metrics $\mathrm{WRSS}_{\text{GNN}}^{(\tau)}$ and $\mathrm{WRSS}_{\text{GBP}}^{(\tau)}$ obtained using \gls{gnn} and \gls{gbp} algorithm, respectively, by $\mathrm{WRSS}_{\text{WLS}}^{(\tau)}$ produced by the \gls{wls} method. For the case when \gls{gbp} or \gls{gnn} algorithms output the same solution as the \gls{wls} method, defined ratio is equal to one, i.e., $(\mathrm{WRSS}_{\text{GBP}}^{(\tau)} / \mathrm{WRSS}_{\text{WLS}}^{(\tau)}) = 1$, $(\mathrm{WRSS}_{\text{GNN}}^{(\tau)} / \mathrm{WRSS}_{\text{WLS}}^{(\tau)}) = 1$.  

\begin{figure}[ht]
	\centering
	\captionsetup[subfigure]{oneside,margin={1.5cm,0cm}}
	\begin{tabular}{@{\hspace{-0.8cm}}c@{\hspace{-0.5cm}}}
	\subfloat[]{\label{plot1a}
	\centering
	\begin{tikzpicture}
        \begin{axis} [box plot width=1.1mm, ymode=log,
    	    xlabel={},
        	ylabel={Normalized WRSS},
  	        grid=major,   		
  	        xmin=0, xmax=3, ymin = 0.7, ymax = 150,	
  	        xtick={1,2},
            xticklabels={GBP,GNN}, 
            ytick={1, 10, 100},
  	        width=4cm,height=4cm,
  	        tick label style={font=\footnotesize}, label style={font=\footnotesize},
  	        legend style={draw=black,fill=white,legend cell align=left,font=\tiny, at={(0.01,0.82)},anchor=west}]
	        \boxplot [
                forget plot, fill=black!20,
                box plot whisker bottom index=1,
                box plot whisker top index=5,
                box plot box bottom index=2,
                box plot box top index=4,
                box plot median index=3] {./plots/gbp30_wrss_1.txt};  
	        \boxplot [
                forget plot, fill=black!20, 
                box plot whisker bottom index=1,
                box plot whisker top index=5,
                box plot box bottom index=2,
                box plot box top index=4,
                box plot median index=3] {./plots/gnn30_wrss.txt};  
	    \end{axis}
	\end{tikzpicture}}
	\end{tabular}\quad\quad\quad
	\begin{tabular}{@{\hspace{-0.2cm}}c@{\hspace{-0.5cm}}}
	\subfloat[]{\label{plot1b}
	\centering
	\begin{tikzpicture}
        \begin{axis} [box plot width=1.1mm, ymode=log,
    	    xlabel={},
        	ylabel={Normalized WRSS},
  	        grid=major,   		
  	        xmin=0, xmax=3, ymin = 60, ymax = 30000,	
  	        xtick={1,2},
            xticklabels={GBP,GNN,}, 
  	        width=4cm,height=4cm,
  	        tick label style={font=\footnotesize}, label style={font=\footnotesize},
  	        legend style={draw=black,fill=white,legend cell align=left,font=\tiny, at={(0.01,0.82)},anchor=west}]
	
	        \boxplot [
                forget plot, fill=black!20,
                box plot whisker bottom index=1,
                box plot whisker top index=5,
                box plot box bottom index=2,
                box plot box top index=4,
                box plot median index=3] {./plots/gbp118_wrss_1.txt};  
	        \boxplot [
                forget plot, fill=black!20, 
                box plot whisker bottom index=1,
                box plot whisker top index=5,
                box plot box bottom index=2,
                box plot box top index=4,
                box plot median index=3] {./plots/gnn118_wrss.txt};  
	    \end{axis}
	\end{tikzpicture}}
	\end{tabular}
	\caption{The normalized WRSS metrics of the GNN and GBP algorithms (we observe WRSS of the GBP algorithm obtained after the first iteration), for power systems with 30 buses (subfigure a) and 118 buses (subfigure b).}
	\label{plot1}
\end{figure}
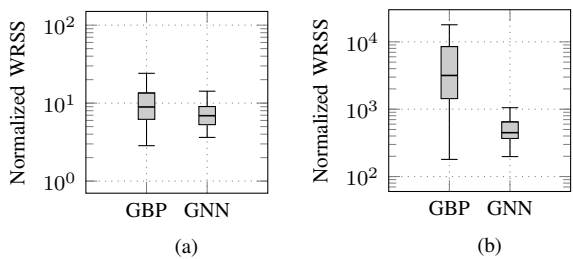

To make a fair comparison, we observe the normalized WRSS metric obtained after the first iteration of the \gls{gbp} algorithm to that of the \gls{gnn} algorithm, when both algorithms provide a solution for the same time complexity. \figurename~\ref{plot1} compares the normalized WRSS metric, $\mathrm{WRSS}_{\text{GBP}}^{(\tau)} / \mathrm{WRSS}_{\text{WLS}}^{(\tau)}$, obtained using the \gls{gbp}, and $\mathrm{WRSS}_{\text{GNN}}^{(\tau)} / \mathrm{WRSS}_{\text{WLS}}^{(\tau)}$ calculated according to the \gls{gnn} algorithm, for each time instance $\tau$ for the power systems with 30 and 118 buses. As shown in \figurename~\ref{plot1}, \gls{gnn} gives better and more consistent results compared to \gls{gbp} in terms of \gls{se} accuracy. This result is intuitive as \gls{gnn} has ability to extract more complex non-linear patterns from data compared to the pure linear \gls{gbp} approach.    

\begin{figure}[ht]
	\centering
	\captionsetup[subfigure]{oneside,margin={1.2cm,0cm}}
	\begin{tabular}{@{\hspace{0cm}}c@{\hspace{0cm}}}
	\subfloat[]{\label{plot2a}
	\centering
	\begin{tikzpicture}
	    \begin{axis} [box plot width=1.1mm, ymode=log,
    	    xlabel={GBP Iteration},
        	ylabel={Normalized WRSS},
  	        grid=major,   		
  	        xmin=0, xmax=10, ymax = 14,	
  	        xtick={1,2,3,4,5,6,7,8,9},
            xticklabels={2,3,4,5,6,7,8,9,10}, 
  	        ytick={1, 10},
  	        width=8cm,height=4cm,
  	        tick label style={font=\footnotesize}, label style={font=\footnotesize},
  	        legend style={draw=black,fill=white,legend cell align=left,font=\tiny, at={(0.01,0.82)},anchor=west}]
	
	        \boxplot [
                forget plot, fill=black!20,
                box plot whisker bottom index=1,
                box plot whisker top index=5,
                box plot box bottom index=2,
                box plot box top index=4,
                box plot median index=3] {./plots/gbp30_wrss_2_10.txt};  
	    \end{axis}
	\end{tikzpicture}}
	\end{tabular}\vspace{3px}
	\begin{tabular}{@{\hspace{0cm}}c@{\hspace{0cm}}}
	\subfloat[]{\label{plot2b}
	\centering
	\begin{tikzpicture}
	    \begin{axis} [box plot width=1.1mm, ymode=log,
    	    xlabel={GBP Iteration},
        	ylabel={Normalized WRSS},
  	        grid=major,   		
  	        xmin=0, xmax=10, ymin = 0.5, ymax = 8000,	
  	        xtick={1,2,3,4,5,6,7,8,9},
            xticklabels={2,3,4,5,6,7,8,9,10}, 
  	        ytick={1, 10, 100, 1000},
  	        width=8cm,height=4cm,
  	        tick label style={font=\footnotesize}, label style={font=\footnotesize},
  	        legend style={draw=black,fill=white,legend cell align=left,font=\tiny, at={(0.01,0.82)},anchor=west}]
	
	        \boxplot [
                forget plot, fill=black!20,
                box plot whisker bottom index=1,
                box plot whisker top index=5,
                box plot box bottom index=2,
                box plot box top index=4,
                box plot median index=3] {./plots/gbp118_wrss_2_10.txt};  
	    \end{axis}
	\end{tikzpicture}}
	\end{tabular}
	\caption{The normalized WRSS metrics of the GBP algorithm during iterations, for power systems with 30 buses (subfigure a) and 118 buses (subfigure b).}
	\label{plot2}
\end{figure}
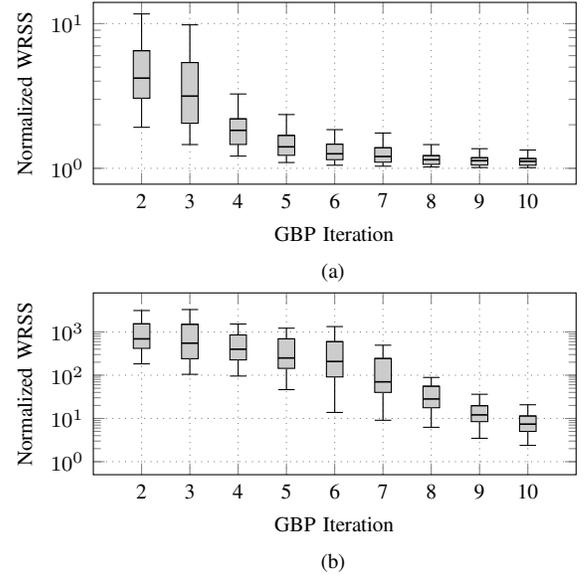

\figurename~\ref{plot2} shows the normalised WRSS metric of the \gls{gbp} algorithm after the first iteration. The \gls{gbp} in a few iterations outperforms the prediction of the \gls{gnn}. Therefore, it can be concluded that the combination of these two algorithms can be a promising solution for fast inference (e.g., \gls{gnn} can provide a more accurate starting point for \gls{gbp}, significantly reducing the total number of iterations), especially since both algorithms are distributed and operate on the same graphical model. 

\subsection{Computation and Communication Delays in SE using 5G}
Critical to establishing a near-real time \gls{se} is the delay introduced by \gls{5g} communication network. Given the \gls{pmu} reporting periods of $10-20\,\text{ms}$, the \gls{se} process needs to produce outputs within the time frame of consecutive reports. At the bottom of \figurename~\ref{Fig_1}, we identify the key delay components for end-to-end \gls{pmu}-to-edge or \gls{pmu}-to-Cloud connectivity. Initial delay is embedded in the \gls{pmu} reporting process and data concentration at \glspl{pdc} \cite{IEEEIECInternationalStandard}. Once the data frame is encapsulated as UDP/IP packet, it is transmitted via \gls{5g} \gls{nr} interface. \gls{5g} \gls{urllc} service is designed to support sub-$1\,\text{ms}$ latency, however, the exact latency the data will experience in disaggregated \gls{oran} system should be carefully examined. In \cite{small2016small,diez2021flexible}, delay components across CU/DU/RU split are investigated, including the delays on fronthaul and midhaul networks. After the UDP/IP packet is received at a nearby \gls{5g} \gls{cn} \gls{upf}, it is routed to a local \gls{sg} edge cloud to distributed \gls{ems} module for processing at the local distributed \gls{se} agent. Communication between neighbouring distributed \gls{se} agents (that run \gls{gbp} or \gls{gnn}-based \gls{se}) proceeds via \gls{5g} \gls{cn} packet delivery between edge cloud servers via one or more \glspl{upf}. Providing precise estimate of delays for \gls{gbp} and \gls{gnn}-based \gls{se} is out of the scope of this paper, but we note that it strongly depends on the mapping of power system factor graph to different edge cloud computation nodes. We note that \gls{gnn} approach may be extremely fast, especially if the $k$-hop neighbourhood nodes are all mapped to the same edge cloud node. In contrast, \gls{gbp} always requires several iterations of message exchanges between neighbouring edge cloud nodes.  

\section{Conclusion}
In this paper, we presented our initial insights on matching recently proposed \gls{ml}/\gls{ai}-based distributed \gls{se} algorithms to the evolving \gls{ml}/\gls{ai}-empowered \gls{5g} network services. Our preliminary investigation indicates the potential of data-based \gls{se} methods to provide near real-time state estimates consistent with \gls{pmu} reporting rates. However, more work is needed, in particular, in the domain of characterisation of complex end-to-end delay statistics, age of information analysis, and overall performance versus delay optimization of distributed \gls{se} functions over \gls{5g} networks.


\bibliographystyle{IEEEtran}
\bibliography{cite}

\end{document}